\documentclass[10pt,twocolumn,letterpaper]{article}

\usepackage{cvpr}            

\definecolor{cvprblue}{rgb}{0.21,0.49,0.74}
\usepackage[pagebackref,breaklinks,colorlinks,allcolors=cvprblue]{hyperref}
\usepackage[sort&compress]{natbib}
\usepackage[utf8]{inputenc} 
\usepackage[T1]{fontenc}   
\usepackage{url}            
\usepackage{booktabs}       
\usepackage{amsfonts}      
\usepackage{nicefrac}       
\usepackage{microtype}      
\usepackage{xcolor}         

\usepackage{natbib}
\setcitestyle{numbers,square}
\usepackage{multirow}
\usepackage{makecell}
\usepackage{amssymb}
\usepackage{bbding}
\usepackage{colortbl}
\usepackage{stfloats}
\usepackage{graphicx}
\usepackage{animate}
\usepackage{amsmath}
\usepackage{caption}
\usepackage{mathtools}
\usepackage{amsthm}
\usepackage[table]{xcolor} 
\usepackage{wrapfig}
\usepackage{float}

\usepackage{capt-of}

\usepackage{orcidlink}

\title{CameraMaster: Unified Camera Semantic-Parameter  Control \\ for Photography Retouching}

\author{
Qirui Yang\textsuperscript{1,2}\thanks{Work was done when Qirui Yang was an intern at vivo.} ~~
Yang Yang\textsuperscript{2} ~~
Ying Zeng\textsuperscript{2} ~~
Xiaobin Hu\textsuperscript{3} ~~ 
Bo Li\textsuperscript{2} ~~
Huanjing Yue\textsuperscript{1} ~~ \\
Jingyu Yang\textsuperscript{1}\thanks{Corresponding author.} ~~
Peng-Tao Jiang\textsuperscript{2}\footnotemark[2] \\
\textsuperscript{1}Tianjin University ~~ \textsuperscript{2}vivo Mobile Communication Co., Ltd ~~ \textsuperscript{3}NUS 
}

\begin{document}

\twocolumn[{
  \renewcommand\twocolumn[1][]{#1}%
  \maketitle
  \begin{center}
  \vspace{-0.7cm}
    \includegraphics[width=0.99\textwidth]{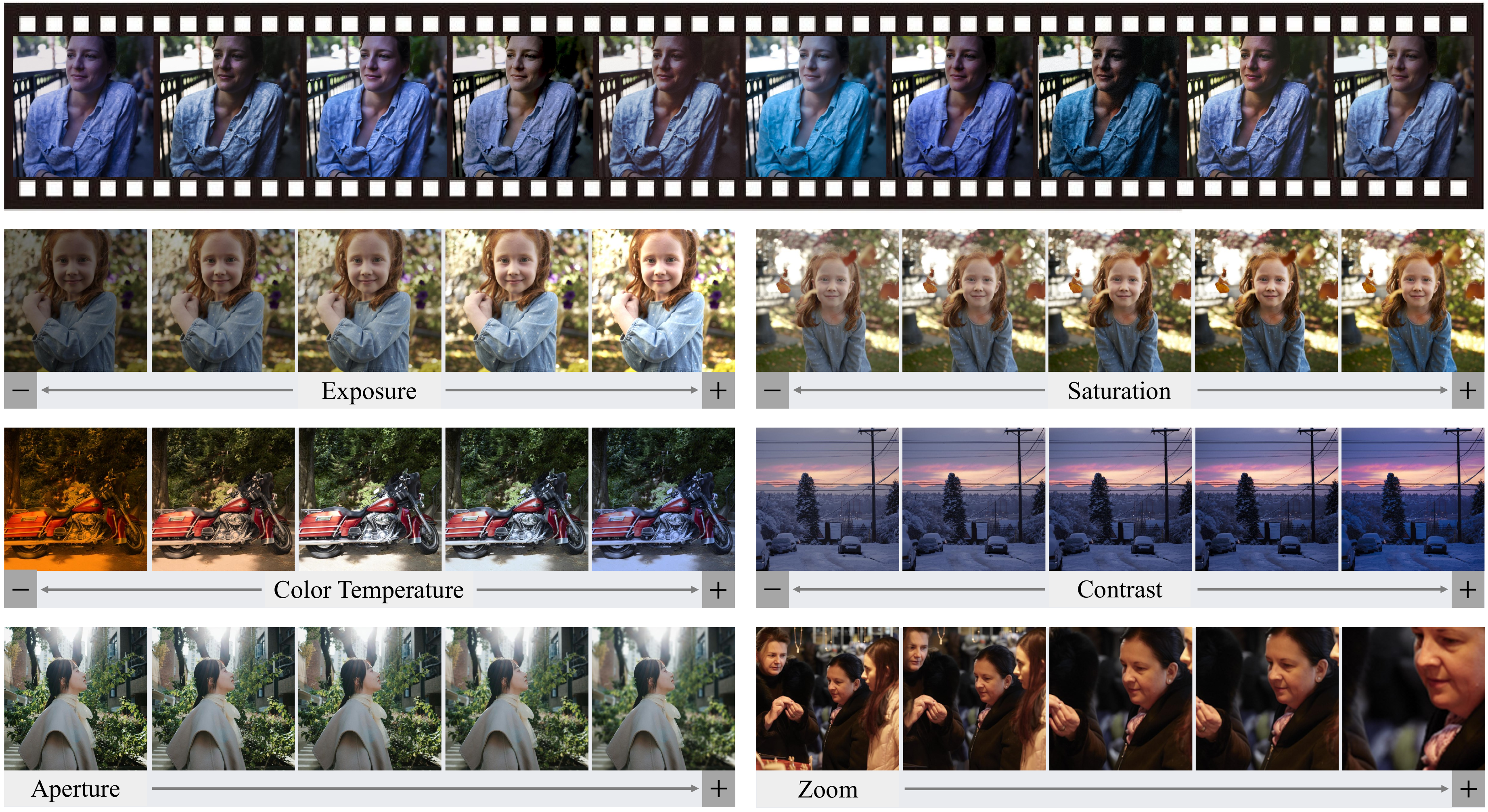}
    \vspace{-0.2cm}
    \captionof{figure}{\textbf{CameraMaster:} Unified, monotonic, and near-linear photography control. Our framework enables precise, physically consistent retouching across seven key camera dimensions, setting a new benchmark for controllable image retouching.
    }
    \label{fig:teaser} 
  \end{center}
}]

\renewcommand{\thefootnote}{\fnsymbol{footnote}}
\setcounter{footnote}{1}
\footnotetext{Work was done when Qirui Yang was an intern at vivo.}

\begingroup
    
\setcounter{footnote}{2}
\footnotetext{Corresponding authors.}
\addtocounter{footnote}{-1}      
\endgroup

\begin{abstract}
Text-guided diffusion models have greatly advanced image editing and generation. However, achieving physically consistent image retouching with precise parameter control (e.g., exposure, white balance, zoom) remains challenging. Existing methods either rely solely on ambiguous and entangled text prompts, which hinders precise camera control, or train separate heads/weights for parameter adjustment, which compromises scalability, multi-parameter composition, and sensitivity to subtle variations. 
To address these limitations, we propose \textbf{CameraMaster}, a unified camera-aware framework for image retouching. The key idea is to explicitly decouple the camera directive and then coherently integrate two critical information streams: a directive representation that captures the photographer’s intent, and a parameter embedding that encodes precise camera settings. CameraMaster first uses the camera parameter embedding to modulate both the camera directive and the content semantics. The modulated directive is then injected into the content features via cross-attention, yielding a strongly camera-sensitive semantic context. 
In addition, the directive and camera embeddings are injected as conditioning and gating signals into the time embedding, enabling unified, layer-wise modulation throughout the denoising process and enforcing tight semantic-parameter alignment. 
To train and evaluate CameraMaster, we construct a large-scale dataset of \textbf{78K} image-prompt pairs annotated with camera parameters. Extensive experiments show that CameraMaster produces monotonic and near-linear responses to parameter variations, supports seamless multi-parameter composition, and significantly outperforms existing methods. 
\end{abstract}
\vspace{-0.1cm}

\vspace{-0.2cm}
\section{Introduction} 
\label{sec:intro}
\vspace{-0.1cm}
\textit{"A photographer must choose his palette like a painter chooses his palette."}—Joel Sternfield

Image retouching \cite{Zhang_llfLUT_2023, su2024styleretoucher, yang2025learning} is fundamental to modern photography. It allows users to manipulate tone, exposure, color temperature, saturation, and other camera-related attributes with precision to enhance emotional expression and atmospheric rendering in photographic works. 
However, achieving such fine-grained control remains highly challenging.

Traditional image retouching methods \cite{Zeng_lut_2020, li2023large, fu2024attentionlut} typically support only one-to-one mapping for fixed tasks and parameters. 
They struggle to jointly adjust multiple camera settings, as shown in Fig. \ref{fig:teaser}, which severely limits style diversity, multi-parameter control, and scene generalization.       
Diffusion-based generative models \cite{rombach2022high, lugmayr2022repaint, chen2024artadapter, zeng2024jedi} and text-guided multimodal large models \cite{zhang2023adding, kumari2023multi} provide powerful text-driven editing for general scenarios, but remain inadequate for precise image retouching.
For directives such as \emph{increase exposure} or \emph{lower color temperature}, these methods often produce ambiguous, non-monotonic responses that may compromise content fidelity.
Overall, both traditional methods and modern generative models lack a precise understanding of camera directives and rigorous modeling of camera parameters. 
This gap has shifted research focus from merely pursuing photorealistic fidelity to achieving predictable, controllable image retouching that explicitly corresponds to camera settings.

Recently, several studies \cite{wang2024humanvid, yuan2025generative, zheng2024cami2v, kuang2024collaborative, bernal2025precisecam} have integrated physical camera signals into generative modeling.
They utilize camera trajectories or viewpoint prompts \cite{xu2024camco, he2025cameractrl} to achieve controllable perspective synthesis, or leverage the bokeh strength in the textual instruction \cite{fortes2025bokeh} to adjust depth-of-field effects.
However, these methods often treat camera parameters as discrete labels or independent regression targets, rather than modeling them as continuous and monotonic control variables in a unified representation space.
For example, \cite{yuan2025generative} adopts differential data, performing differential operations on a set of camera parameter values under the same setting. 
This encourages the model to focus on inter-image differences, yielding smoother and more consistent generation. 
However, constrained by a single setting, such methods usually require separate weights for each parameter configuration.
This design severely compromises scalability and cross-parameter compositionality. 
In summary, existing methods fail to provide continuous, monotonic, and physically consistent responses for joint control of multiple camera parameters in a single unified model.

Consequently, achieving robust, precise, and physically consistent camera control remains challenging for three main reasons:
(i) \textbf{Lack of camera parameter perception.} Current text encoders in diffusion models \cite{yu2024anyedit, zhang2025context} tend to understand camera settings as coarse semantic concepts rather than precise quantitative variables. This leads to editing results that lack continuity, monotonicity, and physical fidelity.
(ii) \textbf{Entanglement between camera directives and content semantics.} Although text encoders \cite{raffel2020exploring, zhai2023sigmoid, flux2024} are powerful, they struggle to cleanly separate camera control intents from mixed prompts. Simply concatenating camera control text with content descriptions often causes brief camera-related semantics to be overwhelmed by strong content semantics.
(iii) \textbf{Absence of a unified, scalable framework.} Existing methods \cite{yuan2025generative, song2025omniconsistency, fortes2025bokeh} typically train separate heads/weights or even independent models for each parameter. This results in fragmented interfaces and reduced sensitivity to parameter combinations and subtle variations.
These limitations motivate us to develop a unified, scalable, and physically consistent framework for camera-aware image retouching.

To address these challenges, we propose \textbf{CameraMaster}, a unified framework for camera-aware image retouching. 
CameraMaster adopts a dual-branch architecture that separately models semantic information and physical camera parameters, realizing parameter-aware and semantic disentanglement, ensuring precise perception and responsive control.
Specifically, we divide the information flow into two branches: a \emph{semantic stream}, which encodes camera directives and content descriptions, and a \emph{parameter stream}, which encodes explicit and quantifiable camera parameters.
To enhance parameter awareness, the parameter stream maps camera parameter values into compact parameter embeddings. 
These parameter embeddings then modulate the semantic stream and time embedding, allowing each component of the model to perceive and respond to precise camera parameters.
To mitigate semantic entanglement, we separately encode the camera directive representation and the content semantic representation, and fuse them through cross-attention, forming a unified semantic feature that is highly sensitive to camera directives.
This unified semantic feature is injected into the DiT backbone to condition generation.
Through separate encoding and interaction between semantic and parameter streams, CameraMaster establishes a robust paradigm for camera-awareness image retouching.
Furthermore, we construct a camera-aware retouching dataset of \textbf{78K} image–prompt pairs annotated with professional camera parameters for model training and evaluation.
Experimental results demonstrate that our method produces monotonic and near-linear responses to parameter variations and achieves superior performance against other baselines in controllable image retouching.
In general, our contributions to the community are threefold:
\begin{itemize}
    \item We propose CameraMaster, a unified framework that achieves monotonic and continuous image retouching through professional camera parameters.
    \item We design a decoupling and fusion mechanism that effectively separates camera parameters from camera directives, ensuring precise perception and responsive control.
    \item We construct a dataset of 78K image–prompt pairs annotated with camera parameters, enabling supervised alignment between semantic and parametric spaces.
\end{itemize}

\vspace{-0.05cm}
\section{Related Work}
\label{sec:related}
\vspace{-0.05cm}
\subsection{Image Retouching and Enhancement}
LUT-based methods model color grading as Look-up Tables \cite{Zeng_lut_2020, fu2024attentionlut, li2024toward, Zhang_llfLUT_2023}, often blending multiple base LUTs with content-dependent weights to approximate professional tone curves \cite{wang2022lcdp, liu2022degradation, wang2022neural, song2021starenhancer, kim2021representative, kim2020global}. 
While efficient and easy to deploy, these methods are largely image-independent, limited in dynamic range, and struggle to realize local, content-aware behaviors, which constrain faithful reproduction of subtle photographic adjustments.
Learning-based methods \cite{yang2022adaint, yang2024taming, yang2024learning} address these issues with end-to-end optimization and structured representations. HDRNet \cite{Gharbi_hdrnet_2017} introduced a bilateral grid representation for fast local tone mapping. Deep 3D LUTs \cite{Zeng_lut_2020} and SepLUT \cite{Yang_SepLUT_2022} learned parameterized color transforms from annotated data. LPTN \cite{Liang_lptn} used Laplacian pyramids to preserve high-frequency detail, and DPRNet \cite{yang2025learning} proposed a differential pyramid representation for high-fidelity retouching.
These methods demonstrate outstanding performance in one-to-one feature mapping, but they only support fixed tasks and parameters, severely limiting their capabilities for multi-task, multi-parameter adjustment, and scene generalization. They lack explicit correlation with multi-camera parameters (such as exposure, white balance, and aperture), prompting research to move beyond fine-grained single retouching toward unified image retouching and editing.

\subsection{Generative and Photographic Image Editing}
Classical editing pipelines relied on explicit pixel operations and handcrafted filters to control geometric or photometric changes \cite{pritch2009shift, barrett2002object, cho2008patch}. They offered precision but limited semantic awareness and contextual adaptability.
Diffusion models enable semantically grounded generation and editing in learned latent spaces \cite{brooks2023instructpix2pix, kawar2023imagic, brack2024ledits++, sheynin2024emu}.
Building on this foundation, recent methods support local, context-preserving edits and extend to mask-based or geometry-aware manipulations \cite{lugmayr2022repaint, rombach2022high, mou2024diffeditor, zhang2023adding, song2025omniconsistency, he2024freeedit, mao2024mag, sajnani2025geodiffuser}.
Instruction-tuned frameworks further adapt diffusion backbones with synthetic instruction–response pairs, and transformer variants improve fidelity and edit consistency on richer data \cite{batifol2025flux, brooks2023instructpix2pix, zhang2025context, xiao2025omnigen}. 
However, these pipelines operate mainly in the semantic domain. Prompts such as “slightly brighter’’ or “warmer tone’’ are intuitive but indirect, often yielding ambiguous, nonlinear, and scene-dependent effects \cite{yuan2025generative, chadebec2025lbm}.
This semantic–physical gap motivates models that respond deterministically and monotonically to camera parameters, bringing generative editing closer to photographic practice.

\subsection{Physically-based Photographic Control}
A complementary line of work integrates physical camera signals into generative modeling. Some works \cite{xu2024camco, he2025cameractrl} leverage camera trajectories or depth prompts for controllable viewpoint synthesis, while others condition on aperture or focal distance for limited depth-of-field effects. Yet, most methods \cite{fortes2025bokeh, yang2025any, yuan2025generative} treat camera variables as discrete tokens or independent regression targets, rather than continuous, interpretable controls within a unified representation.
Moreover, task-specific subnetworks or per-parameter weights, common in current systems for exposure or white balance, undermine scalability, cross-parameter composability, and sensitivity to subtle variations \cite{yuan2025generative, li2024real, chiu2025abc}. 
These limitations point to the need for a single, extensible mechanism that embeds semantic descriptions and numeric magnitudes directly into the generative backbone to achieve stable, monotonic, and disentangled control across photographic factors.
Our work departs from these paradigms by introducing a unified, all-in-one framework that explicitly models the semantic–numeric duality of photographic parameters, thereby aligning the convenience of text-driven editing with the determinism of camera-level control.

\begin{figure*}[t]
    \centering	  
    \vspace{-0.2cm}
    \centering{\includegraphics[width=1\textwidth]{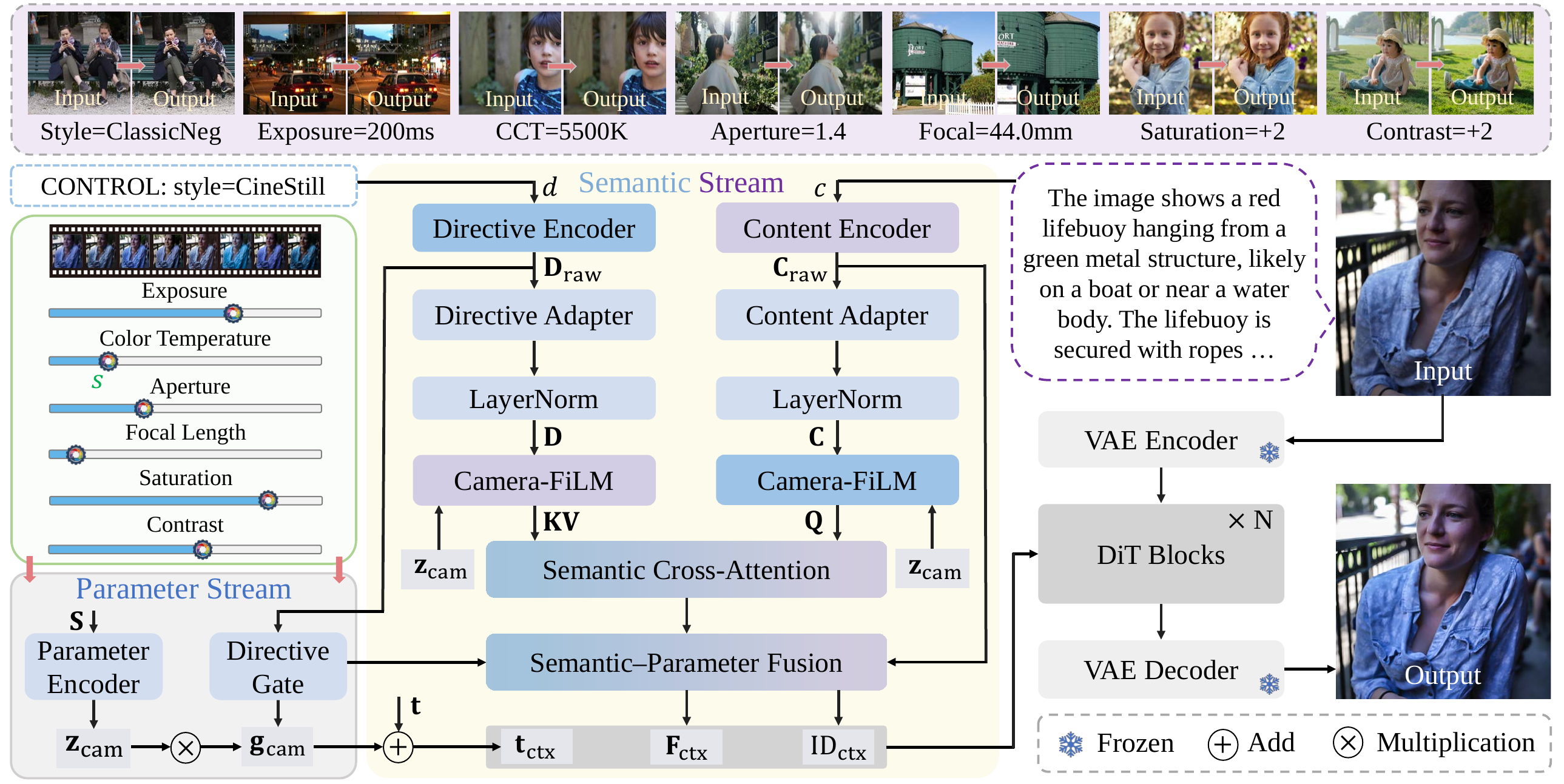}}  
    \vspace{-0.5cm}
    \caption{\textbf{Pipeline of CameraMaster for image retouching.} 
    Given an input image, a camera directive $d$ and a content description $c$, CameraMaster decouples control into two streams: a \textit{semantic stream} that encodes $d$ and $c$, and a \textit{parameter stream} that parses the directive into a calibrated camera vector $s$ and its embedding $\mathbf{z}_{\text{cam}}$.
    The parameter embedding $\mathbf{z}_{\text{cam}}$ modulates both the semantic representations ($\mathbf{C}_{\mathrm{raw}}$, $\mathbf{D}_{\mathrm{raw}}$) and the time embedding $\mathbf{t}$, providing explicit parameter awareness. The modulated features are fused into a unified semantic context and injected into the DiT backbone, enabling continuous, monotonic, and physically consistent camera control for image retouching.
    }
    \vspace{-0.2cm}
    \label{arch}
\end{figure*}

\vspace{-0.1cm}
\section{Method}
\label{sec:method}
\subsection{Overview}
\label{sec:view}
\vspace{-0.1cm}
In this paper, we aim to achieve scalable, robust, and physically consistent camera control for photography retouching. 
As shown in Fig.~\ref{arch}, we propose \textbf{CameraMaster}, a unified framework for camera-aware image retouching built on a diffusion transformer (DiT) backbone.
The model receives three inputs: the input image, an image content description $c$ (caption), and a structured camera control directive $d$ (creative intent).
The core of our framework is how to decouple and interact with semantic information and camera parameters -- containing camera directives, image content, and directive parameters -- to obtain semantic embeddings that are sensitive to camera directives and parameters.
To this end, we adopt a two-branch architecture: a \emph{semantic stream} and a \emph{parameter stream}.
The semantic stream encodes a directive representation from the camera directive $d$ and a content semantic representation from the content description $c$.
These representations are then modulated and fused to produce a unified, camera-aware semantic embedding that serves as the main textual guidance for DiT.
The parameter stream extracts precise numerical values from the camera directive $d$ and encodes them into compact camera parameter embeddings.
The effectiveness of CameraMaster lies in the interaction between these two streams: camera parameter embeddings conditionally modulate the semantic stream, ensuring that the fusion of directives and content is physically grounded and accurately reflects numerical settings.
In addition, the parameter embedding is injected into the DiT backbone by modulating the time embedding.
This dual-path design, where semantic information is physically modulated before guiding the model, while physical parameters provide direct global guidance, enables CameraMaster to achieve robust, continuous, and predictable control.

\vspace{-0.1cm}
\subsection{Decoupled Representations}
\label{sec:decrep}
To avoid semantic ambiguity and ensure precise control, CameraMaster decouples control signals into two independent information streams.
For the semantic stream, we employ two pretrained text encoders \cite{zhai2023sigmoid, raffel2020exploring} to encode guided text into semantic representations.
Given the camera control directive $d$ (e.g., ``[CONTROL: style=CineStill]'') and the image content description $c$, the encoders generate two high-dimensional semantic representations:
\begin{equation}
\mathbf{D}_{\mathrm{raw}} = \mathrm{Encoder}_{\text{SigLIP}}(d), \quad
\mathbf{C}_{\mathrm{raw}} = \mathrm{Encoder}_{\text{T5}}(c), 
\end{equation}
where $\mathbf{D}_{\mathrm{raw}}$ captures the photographer's intent and $\mathbf{C}_{\mathrm{raw}}$ anchors the scene content.
To enable adaptive control strength during subsequent fusion, we further predict a camera directive gate $\mathbf{g}_{\text{cam}} \in (0,1)$ from $\mathbf{D}_{\mathrm{raw}}$ to capture the relative importance of the directive.

In parallel, the parameter stream extracts and quantifies precise camera settings.
To obtain near-linear, monotonic responses, we design a parameter-aware parsing and calibration pipeline that converts free-form values extracted from directives (e.g., ``CineStill'') into physically meaningful normalized scalars.
For example, exposure is represented as a shutter ratio relative to a reference time; color temperature is transformed using logarithmic scaling and normalized; ordinal levels (e.g., contrast) are linearly mapped to $[-1,1]$; and discrete styles are one-hot encoded and then mapped to $[0,1]$.
This calibration step is crucial for achieving physical consistency.
After parsing and normalization, we obtain a calibrated camera parameter vector $\mathbf{s}$. \textit{More details are provided in Section~\ref{sec:dataset}.}

To encode the camera parameter vector into a compact camera parameter embedding, we first broadcast $\mathbf{s}$ into a multi-channel camera parameter tensor $\mathbf{S} \in \mathbb{R}^{B \times 3 \times H \times W}$, whose spatial dimensions match the input image size.
We then introduce a conditional network that encodes the camera parameter tensor into a compact embedding $\mathbf{z}_{\text{cam}}$.
This conditional network consists of four convolutional layers, an adaptive average pooling layer, and an MLP layer.
In this way, we decouple the input signal into two independent, high-dimensional representations: semantic representations ($\mathbf{D}_{\mathrm{raw}}$, $\mathbf{C}_{\mathrm{raw}}$) and a camera parameter embedding ($\mathbf{z}_{\text{cam}}$), which form the basis for subsequent collaborative fusion.

\vspace{-0.1cm}
\subsection{Parameter Modulation and Semantic Fusion}
\label{sec:spcm}
Given the effective decoupling of semantic and parameter information, the next key challenge is their modulation and fusion.
To enable semantic representations to perceive precise camera values and yield monotonic, near-linear responses, we design a camera parameter feature-wise linear modulation, referred to as Camera-FiLM.

To facilitate modulation, we first align the dimensions of the content semantic representation $\mathbf{C}_{\mathrm{raw}}$ and the camera directive representation $\mathbf{D}_{\mathrm{raw}}$ using a camera adapter and a content adapter:
\begin{equation}
\mathbf{D} = \mathrm{Adapter}_{\mathrm{dir}}(\mathbf{D}_{\mathrm{raw}}), \quad
\mathbf{C} = \mathrm{Adapter}_{\mathrm{con}}(\mathbf{C}_{\mathrm{raw}}),
\end{equation}
where each $\mathrm{Adapter}$ module consists of a linear layer, a softmax operation, and a LayerNorm layer.

Camera-FiLM then dynamically generates affine transformation parameters (scaling $\gamma$ and bias $\beta$) from $\mathbf{z}_{\text{cam}}$:
\begin{equation}
\gamma_\text{q}, \beta_\text{q} = \mathrm{FiLM}_\text{q}(\mathbf{z}_{\text{cam}}), \quad
\gamma_\text{kv}, \beta_\text{kv} = \mathrm{FiLM}_\text{kv}(\mathbf{z}_{\text{cam}}).
\end{equation}
These affine parameters modulate the two semantic representations as
\begin{equation}
\mathbf{Q} = (1 + \gamma_\text{q}) \odot \mathbf{C} + \beta_\text{q}, \quad
\mathbf{K}, \mathbf{V} = (1 + \gamma_\text{kv}) \odot \mathbf{D} + \beta_\text{kv}.
\end{equation}

To address semantic ambiguity and prevent the directive signal from being overwhelmed by strong content semantics, we design a semantic cross-attention module that injects the camera directive representation into the content semantic representation:
\begin{equation}
\mathbf{C}_{\text{fuse}} = \mathrm{Softmax} \left( \frac{\mathbf{Q} \mathbf{K}^\top}{\sqrt{d_K}} \right) \mathbf{V},
\end{equation}
where $d_K$ denotes the dimensionality of $\mathbf{K}$.
This encourages the content semantic representation to be highly sensitive to the camera directive representation.

We further introduce an adaptive gating mechanism that modulates the strength of the residual connection using the camera directive gate $\mathbf{g}_{\text{cam}}$:
\begin{equation}
\mathbf{C}_\text{ctx} = \mathbf{C}_{\mathrm{raw}} + \mathbf{g}_{\text{cam}} \cdot \mathbf{C}_{\text{fuse}}.
\end{equation}
The resulting $\mathbf{C}_\text{ctx}$ forms our unified semantic feature, which integrates camera directive intent, content semantics, and camera parameters.

\begin{figure*}[t]
    \centering	  
    \vspace{-0.2cm}
    \centering{\includegraphics[width=1\textwidth]{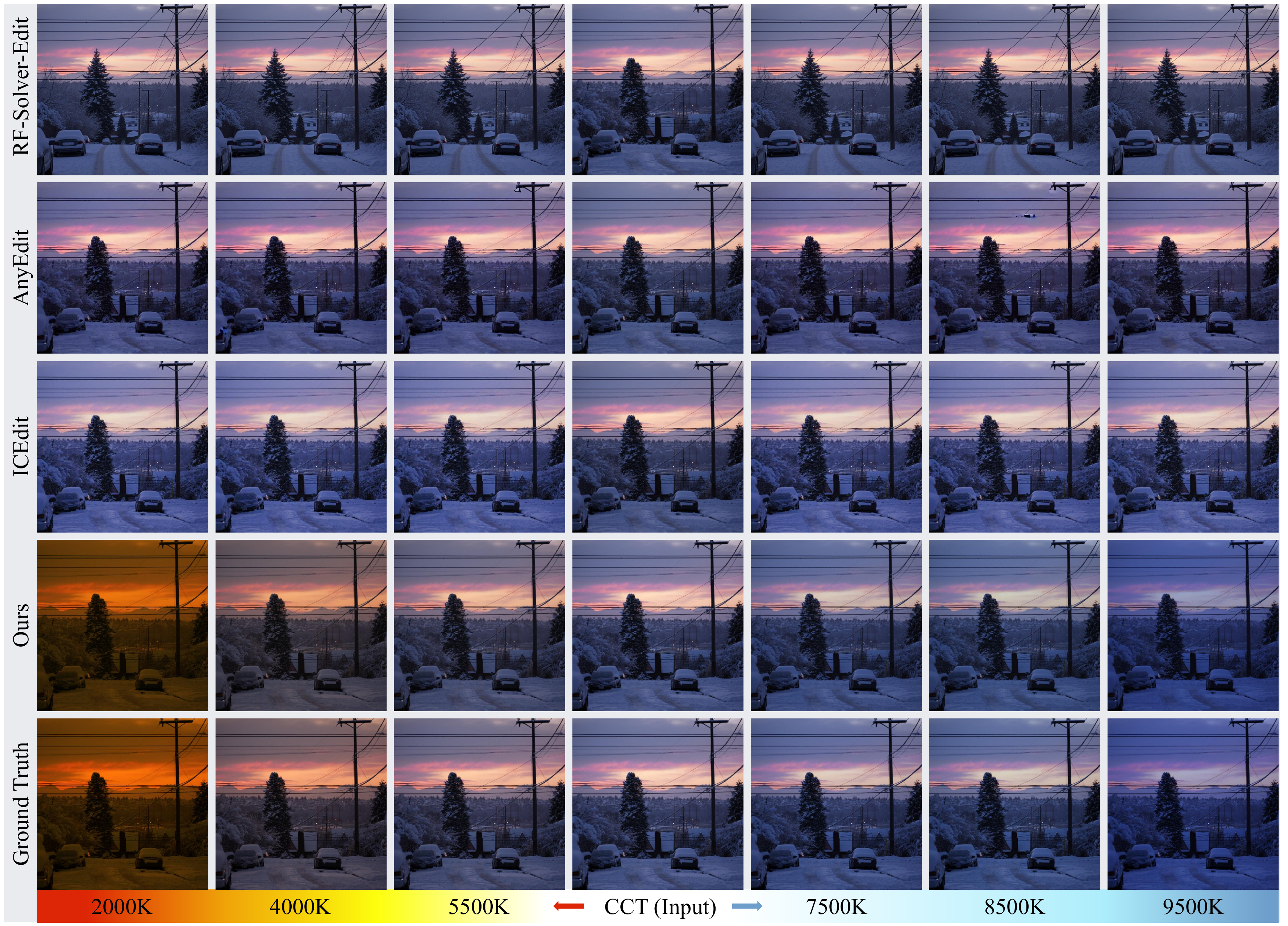}}  
    \vspace{-0.4cm}
    \caption{Visual comparison of Color Temperature (CCT) task retouching results across different methods. This method achieves continuous, monotonous, and precise control over retouching intensity, which strikes a balance between following directions and preserving the original image's consistency to meet user needs.}
    \vspace{-0.2cm}
    \label{fig:cct}
\end{figure*}

\subsection{Semantic-Parameter Information Injection}
\label{sec:inject}
To achieve robust, multi-faceted control, we inject camera information into the DiT through a dual-pathway mechanism. 
This design provides the model with both a rich fused semantic context and a direct physical conditioning signal for global modulation.

Although $\mathbf{C}_{\text{ctx}}$ provides a holistic semantic context, the fusion process can dilute the salience of individual control directives across many tokens.
To preserve explicit control, we complement the fused context with an addressable camera directive representation.
Specifically, we compress the original camera directive representation $\mathbf{D}$ with two linear layers to obtain a compact camera directive context $\mathbf{D}_{\text{dir}}$.
We then concatenate this with the fused context to form the full semantic input $\mathbf{F}_{\text{ctx}}$:
\begin{equation}
\mathbf{F}_{\text{ctx}} = \mathrm{Concat}(\mathbf{C}_{\text{ctx}}, \mathbf{g}_{\text{cam}} \cdot \mathbf{D}_{\text{dir}}),
\end{equation}
where the camera directive gate $\mathbf{g}_{\text{cam}}$ controls the contribution of the explicit directive.
By providing both the blended context $\mathbf{C}_{\text{ctx}}$ and the camera directive context $\mathbf{D}_{\text{dir}}$, we enable the DiT's attention mechanism to learn the overall camera control.

To further enhance fine-grained control, we provide explicit positional information by concatenating the positional encodings for the content and the camera directive:
\begin{equation}
\mathbf{ID}_{\text{ctx}} = \mathrm{Concat}(\mathbf{ID}_{\text{content}}, \mathbf{ID}_{\text{dir}}),
\end{equation}
where $\mathbf{ID}_{\text{content}}$ denotes the standard sequential indices for the content tokens and $\mathbf{ID}_{\text{dir}}$ is assigned a new block of continuous integer indices that follow the last content index.
This creates a unified, unambiguous positional map for the entire input sequence.

In this way, the DiT learns not only what the control is, but also where its semantic representation is located, enabling more precise and disentangled retouching.
Furthermore, the camera parameter embedding $\mathbf{z}_{\text{cam}}$ modulates the time embedding $\mathbf{t}$:
\begin{equation}
\mathbf{t}_\text{ctx} = \mathbf{t} + \mathbf{g}_{\text{cam}} \cdot \psi(\mathbf{z}_{\text{cam}}),
\end{equation}
where $\psi$ is a two-layer MLP.
The modulated embedding $\mathbf{t}_\text{ctx}$ conditions all AdaLN layers within the DiT blocks.
Through this architecture, CameraMaster closely aligns semantic intentions with physical parameter values, achieving precise, composable, and physically coherent control.

\section{Experiments}
\label{exp}

\subsection{Dataset: CameraSet}
\label{sec:dataset}

\noindent\textbf{Goal and Scope.}
To support monotonic, continuous, and fine-grained photo retouching, we construct a large-scale dataset, \textbf{CameraSet}, specifically designed for real-world camera-aware retouching.
CameraSet selects 2K images from the FilmSet dataset \cite{li2023large} as base images, with wide dynamic range, appropriate exposure, and neutral color temperature.
On top of these, we physically simulate multiple camera functions, including film style, color temperature, exposure, zoom, contrast, saturation, and bokeh.
Each function comprises multiple quantifiable parameter settings analogous to real camera controls.

We categorize CameraSet into several calibrated settings:
(i) \emph{Film Simulation}: To support emotional expression and atmospheric rendering, we employ multiple photographers to meticulously simulate various classic film styles. \textbf{10} iconic films (such as ClassicNeg, Velvia, Kodak Gold, etc.) are selected to offer diverse options.
(ii) \emph{Exposure}: We adjust the irradiance of the base image and convert it back to RGB using a simulated image signal processor (ISP) \cite{li2024efficient, yuan2025generative}, producing images with different exposure levels.
(iii) \emph{Color Temperature}: We use an empirical approximation \cite{fairchild2013color} based on the relationship between blackbody radiation and color temperature to adjust RGB channel ratios for different color temperatures.
(iv) \emph{Contrast and Saturation}: We synthesize four levels of contrast and saturation using Photoshop, covering common retouching ranges.
(v) \emph{Zoom}: We follow end-to-end methods \cite{mallat2002theory, witkin1987scale, yuan2025generative}. For each base image, we compute the field-of-view (FoV) ratio corresponding to a sampled focal length relative to the original focal length.
(vi) \emph{Bokeh}: We construct 400 base–foreground pairs and render four levels of blur to simulate different apertures and subject distances \cite{yang2025any}.

All values are normalized into a standard camera vector $\mathbf{s}$ (see Section~\ref{sec:decrep}), ensuring that equal text strides correspond to equal numerical strides across different control parameters, consistent with the pre-attention conditioning setup.

\noindent\textbf{Pair Generation and Calibration.}
For each base image, we generate input–output pairs by applying physically driven parametric operations.
Each data point is annotated with: (a) a \emph{camera control directive} $d$, and (b) a content description $c$ (image caption) describing the image, generated by a vision–language model \cite{bai2025qwen2}.
This pairing aligns semantic intent with numerical targets and forms the basis for evaluating monotonic control curves.

\noindent\textbf{Dataset Details} 
CameraSet provides 2,000 training pairs per control parameter, for a total of \textbf{78K} training pairs.
For evaluation, CameraSet includes \textbf{570} test pairs.
The test set is disjoint from the training set and covers all control parameters, enabling assessment of continuity, monotonicity, and identity preservation under controlled editing conditions.
To further evaluate CameraMaster’s generalization in real-world scenarios, we also collect an additional dataset of real scenes without paired ground truth.

\subsection{Implementation Details}
\label{sec:details}
We train our model using the AdamW optimizer with a learning rate of $5\times 10^{-4}$.
The training resolution is set to $1024{\times}1024$.
All experiments are conducted on 8 NVIDIA H20 GPUs.
To comprehensively evaluate our model, we compute PSNR, SSIM, LPIPS \cite{zhang2018unreasonable}, PDist, $\Delta E$ ~\cite{zhang1996spatial}, CLIP-I \cite{hessel2021clipscore, shafiullah2022clip}, DINO \cite{caron2021emerging}, and FID \cite{heusel2017gans} metrics to measure the discrepancy between model outputs and ground truth.

\begin{figure*}[t]
    \centering	  
    \centering{\includegraphics[width=\textwidth]{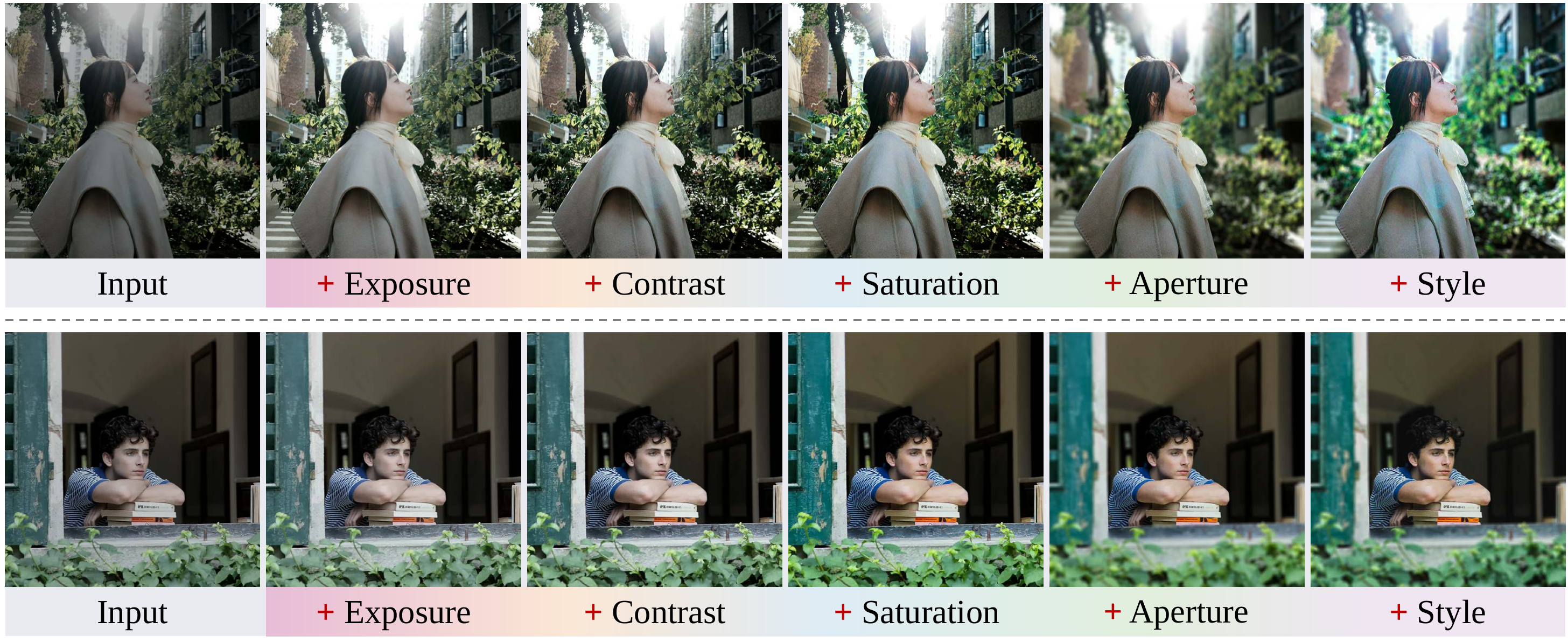}}  
    \vspace{-0.5cm}
    \caption{Examples showcase the cross-parameter retouching power of CameraMaster on the real-world dataset. Our method allows for continuous and fine-grained retouching across multiple parameter combinations, illustrating CameraMaster's universal capability in achieving diverse retouching objectives.}
    \vspace{-0.1cm}
    \label{mulstep}
\end{figure*}

\begin{table*}[t]
  \centering
  \caption{\textbf{Quantitative results on CameraSet test set.} All metrics are calculated between the retouching image and the ground truth image provided by the CameraSet.}
  \label{tab:CameraSet}
  \vspace{-3mm}
  \small
  \renewcommand\arraystretch{0.83}
  \resizebox{0.8\linewidth}{!}{%
  \begin{tabular}{ccccccccc}
    \toprule[1pt]
    \rowcolor[HTML]{FAFAFA}
    \multicolumn{1}{c}{Methods} & PSNR$\uparrow$ & SSIM$\uparrow$ & LPIPS$\downarrow$ & PDist $\downarrow$ & $\Delta \mathbf{E}$ $\downarrow$ & $\text{CLIP-I}\uparrow$ & $\text{DINO}\uparrow$ & FID$\downarrow$ \\
    \midrule
    \rowcolor[HTML]{F9FFF9}
    RF-Solver-Edit*\hspace{0.2em} \cite{wang2024taming}      & 19.68   & 0.7332 & 0.3556 & 15.03 & 10.38  &0.8719  & 0.8354 & 93.64\\
    \rowcolor[HTML]{F9FFF9}
    FLUX.1 Fill*\hspace{0.2em} \cite{flux2024}   &10.19   &0.3708  & 0.7733 & 23.03 &25.90   &0.6480  &0.4836  & 202.53\\ 
    \rowcolor[HTML]{F9FFF9}
    FLUX.1 Kontext*\hspace{0.2em} \cite{labs2025flux1kontextflowmatching}  &11.91   &0.4262  &0.6807  &22.42  & 23.93  &0.7826  &0.7014  &115.48 \\ 
    \rowcolor[HTML]{F9FFF9}
    InstructP2P\hspace{0.2em} \cite{brooks2023instructpix2pix}  & \underline{26.47}  & 0.8002 & \underline{0.1319} & 12.96  & \underline{4.40}  &\underline{0.9643}  & \underline{0.9478} & \underline{30.46} \\ 
    \rowcolor[HTML]{F9FFF9}
    OminiControl\hspace{0.2em} \cite{tan2025ominicontrol}  & 14.72  & 0.6026  & 0.5241  & 17.90 & 16.18   & 0.7533 & 0.6921  & 140.25\\ 
    \rowcolor[HTML]{F9FFF9}
    AnyEdit\hspace{0.2em} \cite{yu2024anyedit}  & 25.48  & 0.7871  & 0.1572 & 12.44 & 4.89  & 0.9383 & 0.8854 & 37.63 \\ 
    \rowcolor[HTML]{F9FFF9}
    ICEdit\hspace{0.2em} \cite{zhang2025context}  & 23.57   & \underline{0.8079} & 0.2209 & \underline{10.91} & 9.15  & 0.9266 & 0.9370 & 31.50\\ 
    \midrule
    \rowcolor[HTML]{F9FBFF}
    \multicolumn{1}{c}{\ \ \ \ \textbf{CameraMaster}}   & \textbf{32.80}   & \textbf{0.9240} & \textbf{0.0669} & \textbf{5.66} &  \textbf{3.07} & \textbf{0.9846} & \textbf{0.9865} & \textbf{7.70} \\
    \bottomrule[1pt]
  \end{tabular}}
  \vspace{-4.0mm}
\end{table*}

\subsection{Comparisons with State-of-the-Art}

\noindent\textbf{Quantitative Comparison.} To evaluate CameraMaster's performance, we quantitatively compared it against a suite of state-of-the-art image editing methods on our constructed CameraSet test dataset. The benchmark methods span multiple paradigms, including off-the-shelf powerful training-free universal editors (RF-Solver-Edit \cite{wang2024taming}, FLUX.1 \cite{flux2024}, FLUX.1 Kontext \cite{labs2025flux1kontextflowmatching}) and retrained instruction-driven editing models (InstructP2P \cite{brooks2023instructpix2pix}, OminiControl \cite{tan2025ominicontrol}, AnyEdit \cite{yu2024anyedit}, ICEdit \cite{zhang2025context}).
As shown in Tab.~\ref{tab:CameraSet}, CameraMaster significantly outperforms all baseline methods across all 8 evaluation metrics, achieving state-of-the-art performance. This demonstrates the effectiveness of our framework for precise, physically consistent image retouching.
In particular, CameraMaster shows clear advantages in reconstruction fidelity and perceptual quality.
CameraMaster achieves 32.80 dB in terms of PSNR metric, surpassing the strongest baseline method, InstructP2P, by 6.33 dB.
Moreover, CameraMaster's LPIPS score is only 0.0669. Regarding color accuracy, the $ \Delta \mathbf{E}$ metric reached 3.07, significantly outperforming all competitors. 
This success stems directly from our decoupled semantic and parameter stream, which provides precise numerical guidance to the model, eliminating the ambiguity inherent in pure text instructions. 
Furthermore, in terms of content and semantic consistency, the CLIP-I and DINO scores (0.9846 and 0.9865, respectively) both rank first. 
This demonstrates that CameraMaster maximally preserves the original image's content and identity characteristics while executing precise camera parameter adjustments, effectively avoiding the “content drift” or identity destruction issues common in other methods. 

To further validate and evaluate CmaeraMaster, especially in more challenging real-world scenarios lacking ground truth benchmarks, we have conducted additional experiments. As shown in Tab.~\ref{tab:realworld}, our model achieves state-of-the-art performance across all relevant metrics, including the lowest CLIP-DIST and DINO-DIST, as well as the highest CLIP-SIM and DINO-SIM.
The achievement stems from our parameter modulation and semantic fusion mechanism.
Finally, CameraMaster achieved a decisive victory in the photorealism and distributional quality of generated images. 
In summary, quantitative comparisons unequivocally validate the design advantages of the CameraMaster framework. 
Compared to generic editing models reliant on ambiguous semantic instructions, our approach successfully establishes a new technical benchmark across all critical dimensions of image retouching—precision, fidelity, content preservation, and photorealism—by decoupling semantic intent from camera parameters and efficiently aligning them.

\begin{table}[t]
	\centering
	\caption{\textbf{Quantitative results on real-world dataset.}}
    \label{tab:realworld}
	\vspace{-3.5mm}
\centering
\small
\renewcommand\arraystretch{1.1}
\resizebox{\linewidth}{!}{
\begin{tabular}{ccccc}
\bottomrule[1pt]\rowcolor[HTML]{FAFAFA}
\multicolumn{1}{c}{Settings}   & CLIP-DIST$\uparrow$ & CLIP-SIM$\uparrow$ & DINO-DIST$\downarrow$ & DINO-SIM $\uparrow$\\ 
\toprule[0.8pt]
\rowcolor[HTML]{F9FFF9}
RF-Solver-Edit   &0.0889   &0.9111  &0.1154  &0.8846           \\ 
\rowcolor[HTML]{F9FFF9} FLUX.1 Fill  &0.2969   &0.7031  &0.4755  & 0.5245         \\
\rowcolor[HTML]{F9FFF9} InstructP2P   &0.0971   &0.9029  & 0.0725 &0.9275 \\
\rowcolor[HTML]{F9FFF9} OminiControl   &0.2174   &0.7827  & 0.1860 &0.8140 \\
\rowcolor[HTML]{F9FFF9} AnyEdit   &0.0751   &0.9248  & 0.0623 &0.9377 \\
\rowcolor[HTML]{F9FFF9} ICEdit   &0.0674   &0.9325  & 0.0363 &0.9636 \\
\midrule[0.8pt]
\rowcolor[HTML]{F9FBFF} \multicolumn{1}{c}{\ \ \ \
 Ours  } & \textbf{0.0293}   & \textbf{0.9707} & \textbf{0.0184} & \textbf{0.9816}   \\
\toprule[1pt]
\end{tabular}}
\vspace{-4.0mm}
\end{table}

\noindent\textbf{Qualitative Results.}
As shown in Fig.~\ref{fig:cct}, the results show a qualitative comparison on the color temperature (CCT) task. 
When we gradually vary the CCT value, CameraMaster produces smooth and monotonic changes in global color tone, while preserving scene structure and local details. 
In contrast, diffusion-based baselines often exhibit non-monotonic shifts and or failures. 
These results confirm that semantic-parameter alignment enables precise, continuous control of retouching strength and maintains a good balance between following camera directives and keeping the original content intact.
As shown in Fig.~\ref{mulstep}, CameraMaster operates on real images with joint control of multiple parameters (e.g., exposure, saturation, style). 
Our method supports fine-grained, continuous transitions across different parameter combinations, yielding coherent and natural retouching results while preserving identity and scene layout. 
Compared with baselines, CameraMaster better respects each target parameter dimension and avoids unintended changes, highlighting its generality and practicality for diverse, camera-centric retouching goals. \textit{More qualitative results can be found in the supplementary materials.}

\subsection{Ablation Study}
\label{sec:ablation}
We conducted ablation experiments to validate the effectiveness of key components in our CameraMaster framework. 
All experiments were performed on the CameraSet test dataset, with results shown in Tab.~\ref{tab:ablation}.
We first established a baseline model that removed our decoupled design, directly feeding text prompts with camera parameters into the text encoder.
Concurrently, we included an off-the-shelf benchmark \cite{labs2025flux1kontextflowmatching} to demonstrate the necessity of fine-tuning. 
Our complete model significantly outperforms the baseline across all key metrics. 
Particularly on the PSNR metric, our model achieves 32.80 dB, surpassing the baseline (28.54 dB) by 4.26 dB. 
This strongly demonstrates that simply concatenating semantic and parametric information in text is a suboptimal strategy, while our proposed architecture more effectively leverages both types of information to achieve higher-fidelity image retouching.

In the \textit{w/o Parametric setting}, we removed the camera parameter stream. 
Experimental results show PSNR significantly dropped from 32.88 dB to 30.68 dB. This substantial performance degradation indicates that camera parameter embeddings are crucial for achieving pixel-level precise reconstruction. 
In the \textit{w/o Semantic setting}, we removed the semantic stream, allowing the model to receive only calibrated physical parameter vectors without knowing which specific camera parameters these values correspond to. 
Under these conditions, the model's performance also deteriorated across the board, particularly in perceptual metrics LPIPS (degrading from 0.0669 to 0.0856) and color accuracy $\Delta E$ (worsening from 3.073 to 3.902). 
This demonstrates that camera instruction representations provide indispensable contextual information for camera parameters. Without semantic instruction guidance, the model struggles to correctly interpret and apply these numerical values, leading to perceptual and color deviations in the generated results.
An intriguing phenomenon is that despite the baseline model's overall poor performance, its $\Delta E$ metric is unusually good. Our analysis suggests this occurs because it fails to effectively interpret complex editing instructions, instead tending to execute very conservative, minimal edits. 
While this failure mode causes minimal color disturbance, it completely fails to accomplish the intended retouching task, as evidenced by its extremely low PSNR values.

\begin{table}[t]
	\centering
	\caption{\textbf{Ablation study on model structure.} We evaluate the performance of different settings on the CameraSet test set.}
    \label{tab:ablation}
	\vspace{-3.5mm}
\centering
\small
\renewcommand\arraystretch{1.1}
\resizebox{0.9\linewidth}{!}{
\begin{tabular}{ccccc}
\bottomrule[1pt]\rowcolor[HTML]{FAFAFA}
\multicolumn{1}{c}{Settings}   & PSNR$\uparrow$ & SSIM$\uparrow$ & LPIPS$\downarrow$ & $\Delta \mathbf{E}$ $\downarrow$\\ 
\toprule[0.8pt]
\rowcolor[HTML]{F9FFF9}
Training-free   &11.91   &0.4262  &0.6807  &23.93           \\ 
\rowcolor[HTML]{F9FFF9} Benchmark  &28.54   &0.9203  &0.0599  & 0.981         \\
\rowcolor[HTML]{F9FFF9} w/o Parametric   &30.68   &0.9237  & 0.0541 &3.182 \\
\rowcolor[HTML]{F9FFF9} w/o Semantic  &31.27   &0.9145  & 0.0856 & 3.902         \\
\midrule[0.8pt]
\rowcolor[HTML]{F9FBFF} \multicolumn{1}{c}{\ \ \ \
 Full model  } & \textbf{32.80}   & \textbf{0.9240} & \textbf{0.0669} & \textbf{3.073}   \\
\toprule[1pt]
\end{tabular}}
\vspace{-4.0mm}
\end{table}

\section{Conclusion}
\label{conc}
In this paper, we introduce \textbf{CameraMaster}, a unified framework for camera-aware photographic retouching. 
Our key idea is to explicitly separate control signals into two information streams: a stream for photographic instruction representations and a stream for camera parameter embeddings that encode precise numerical settings. 
Through tailored modulation, cross-attention, and unified injection mechanisms, CameraMaster effectively aligns and fuses these two streams within a single diffusion-based model.
We conduct extensive experiments on a newly constructed dataset of 78K image–prompt pairs with camera parameter annotations. 
Experimental results demonstrate that our method produces monotonic and near-linear responses to parameter variations and achieves superior performance against other baselines in controllable image retouching.

{
    \small
    \bibliography{main}
}

\clearpage
\setcounter{page}{1}
\maketitlesupplementary

\section{More Qualitative Results}
\begin{figure*}[t]
    \centering	  
    \vspace{-0.2cm}
    \centering{\includegraphics[width=1\textwidth]{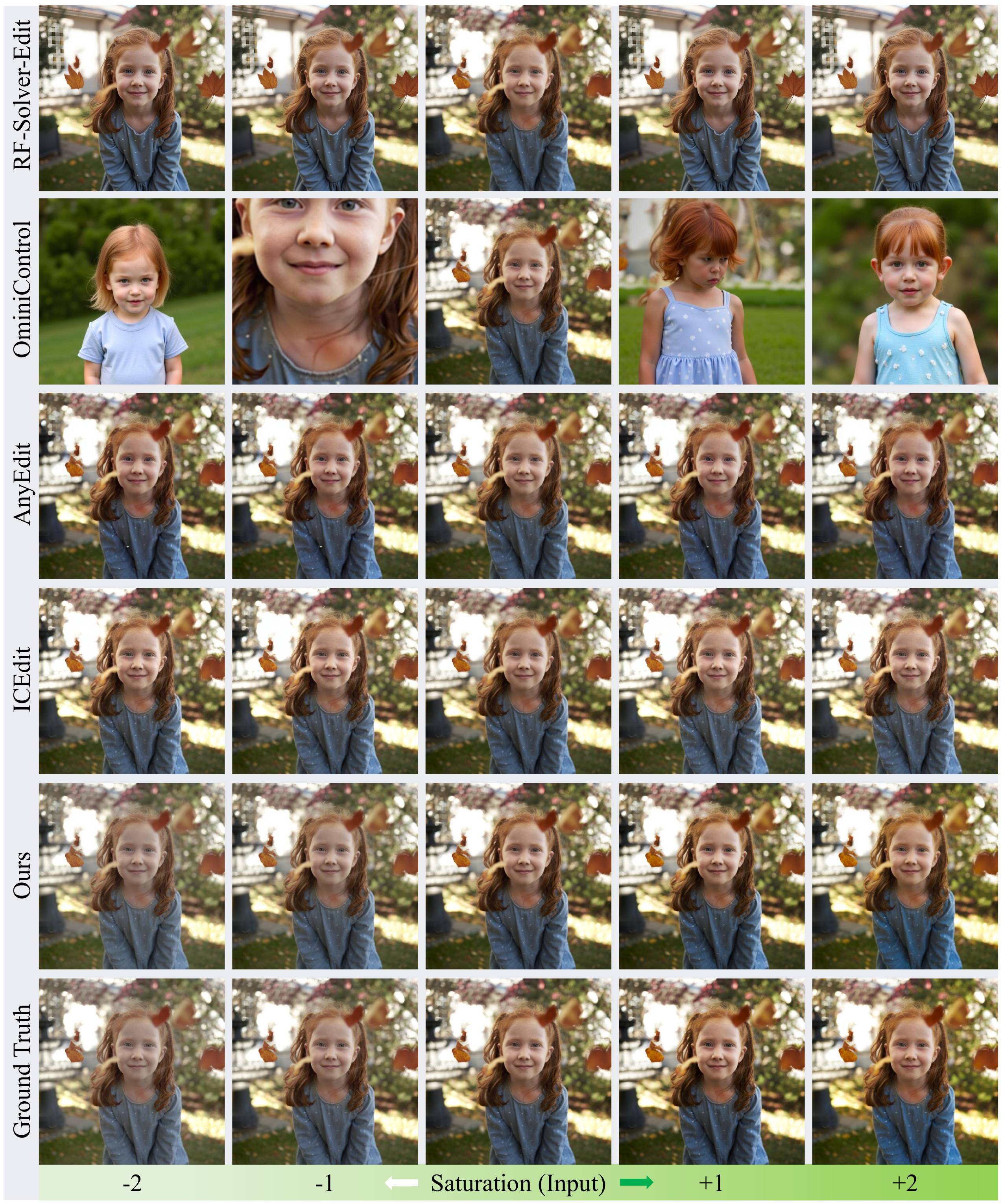}}  
    \vspace{-0.5cm}
    \caption{Visual comparison of \textbf{Saturation} task retouching results across different methods.}
    \vspace{-0.3cm}
    \label{fig:saturation}
\end{figure*}

\begin{figure*}[t]
    \centering	  
    \vspace{-0.2cm}
    \centering{\includegraphics[width=1\textwidth]{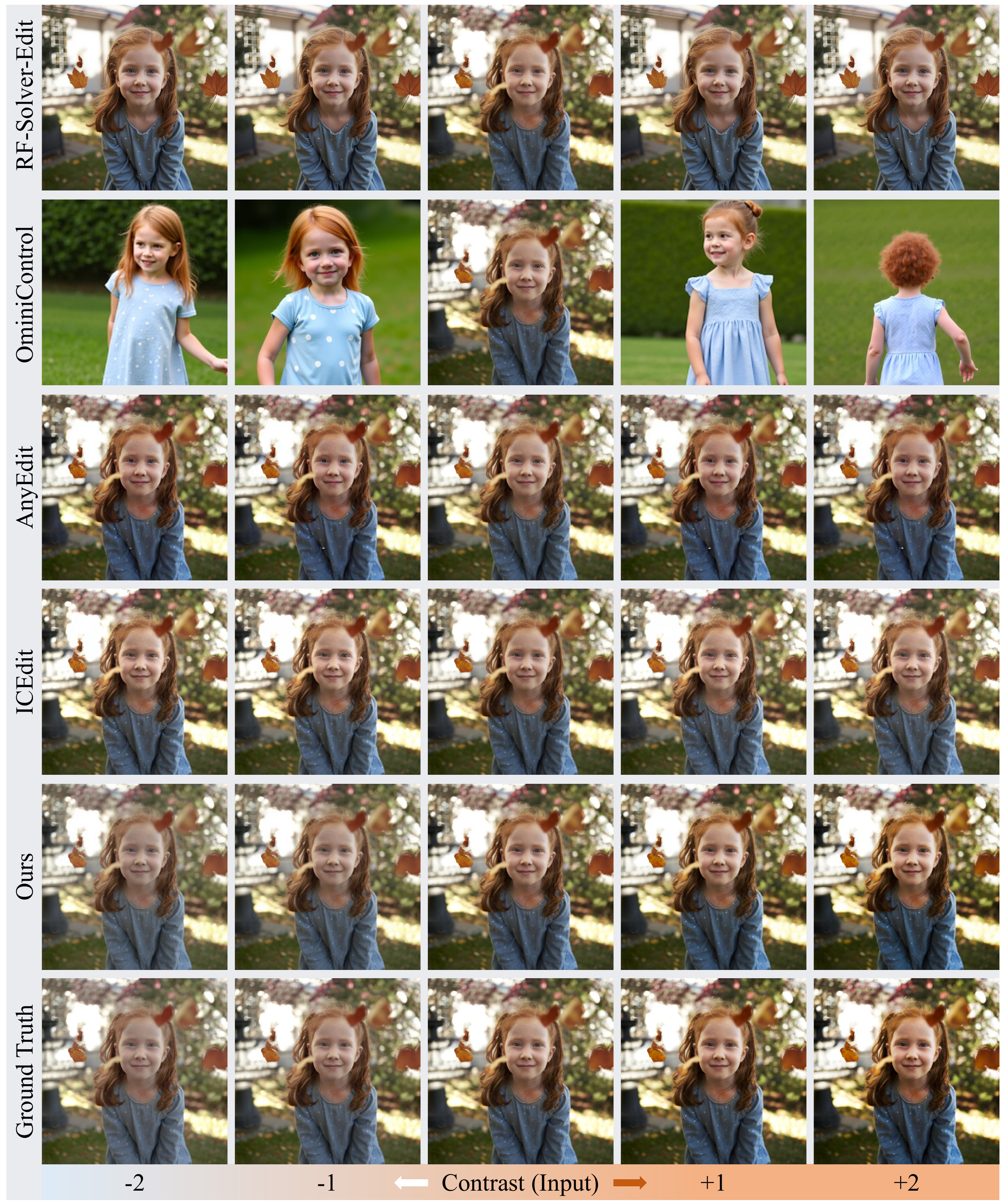}}  
    \vspace{-0.5cm}
    \caption{Visual comparison of \textbf{Contrast} task retouching results across different methods.}
    \vspace{-0.3cm}
    \label{fig:contrast}
\end{figure*}

\begin{figure*}[t]
    \centering	  
    \vspace{-0.2cm}
    \centering{\includegraphics[width=1\textwidth]{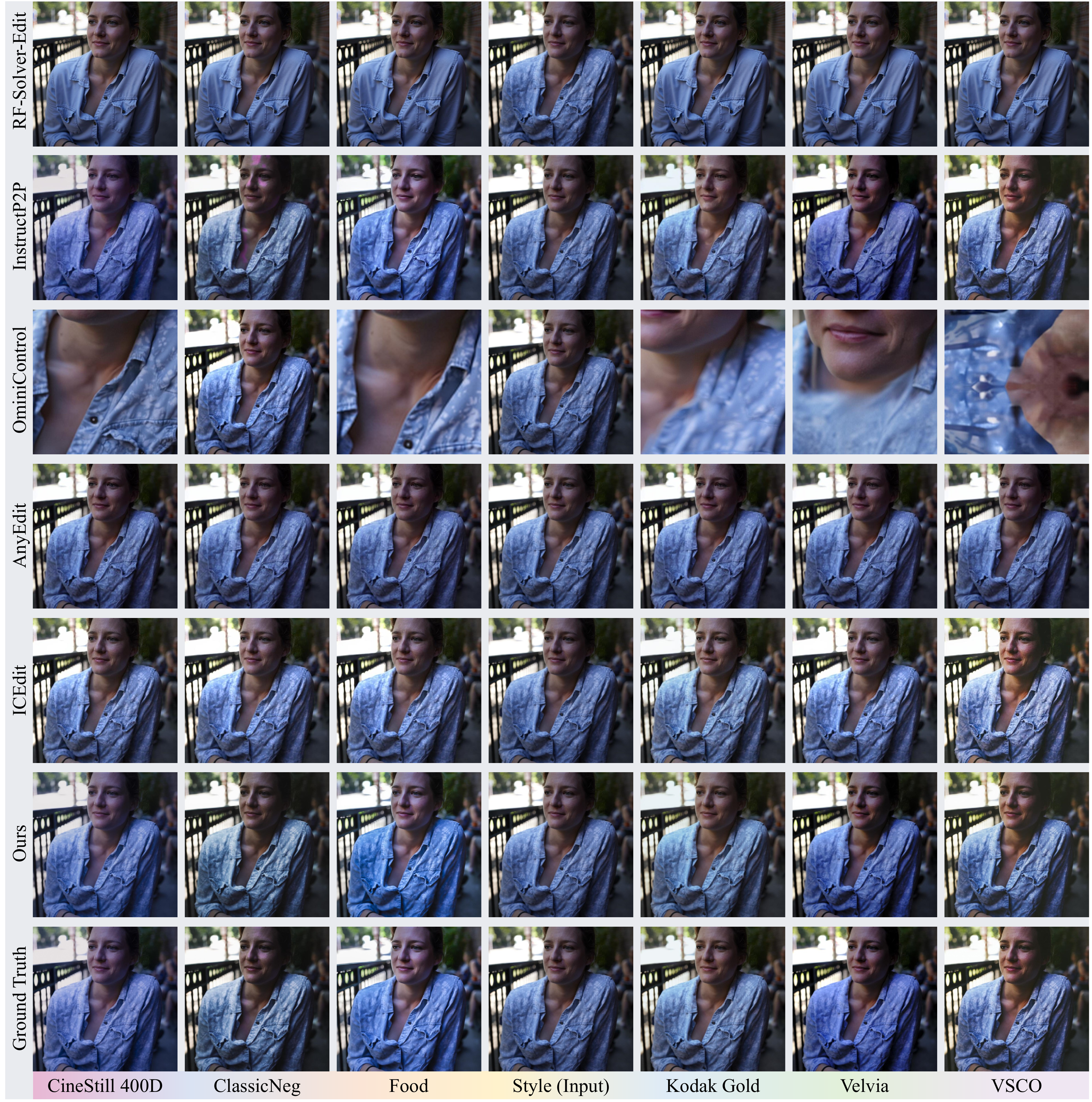}}  
    \vspace{-0.5cm}
    \caption{Visual comparison of \textbf{Style} task retouching results across different methods.}
    \vspace{-0.3cm}
    \label{fig:style}
\end{figure*}

\begin{figure*}[t]
    \centering	  
    \vspace{-0.2cm}
    \centering{\includegraphics[width=1\textwidth]{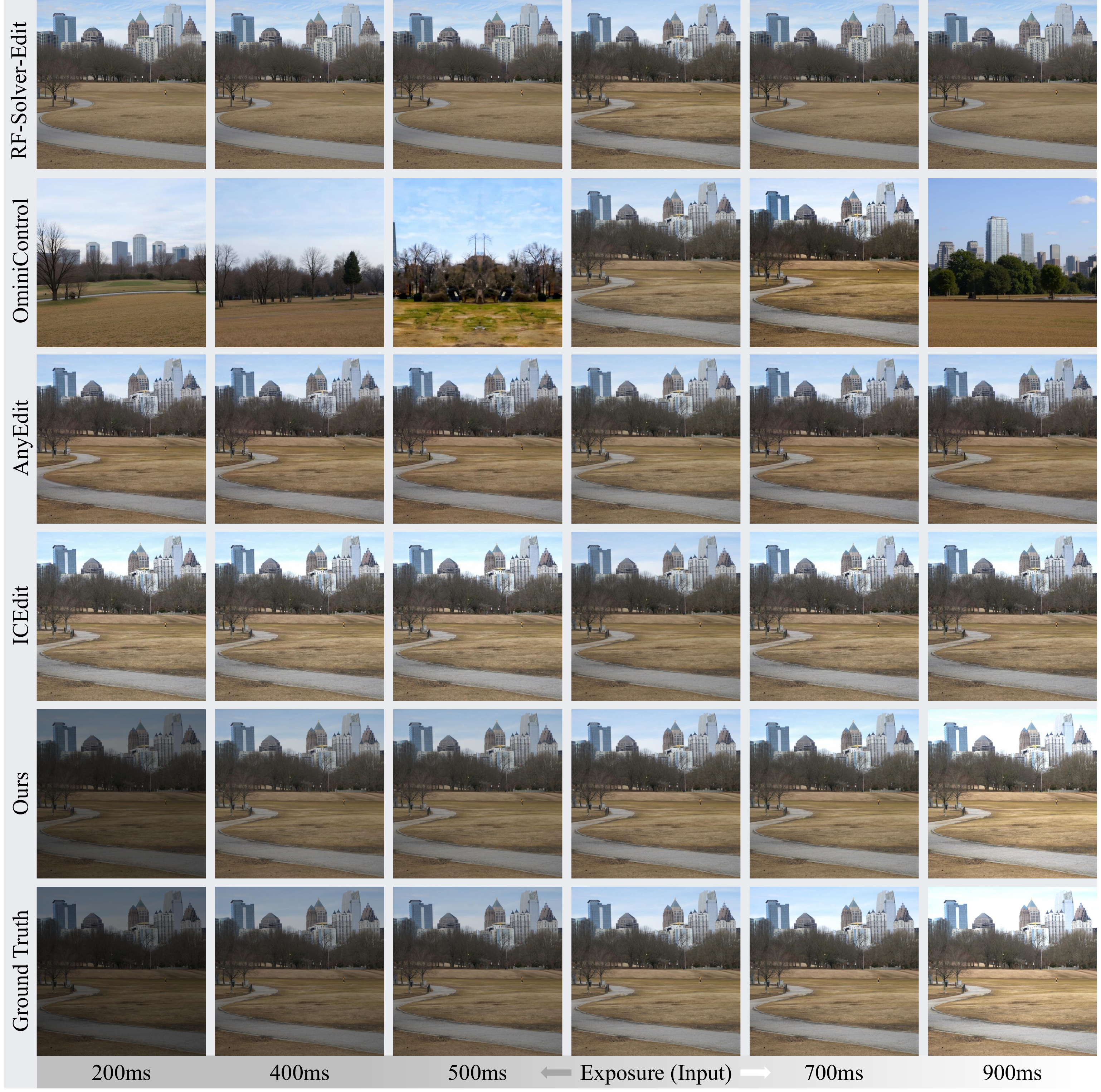}}  
    \vspace{-0.5cm}
    \caption{Visual comparison of \textbf{Exposure} task retouching results across different methods.}
    \vspace{-0.3cm}
    \label{fig:expo}
\end{figure*}

\begin{figure*}[t]
    \centering	  
    \vspace{-0.2cm}
    \centering{\includegraphics[width=1\textwidth]{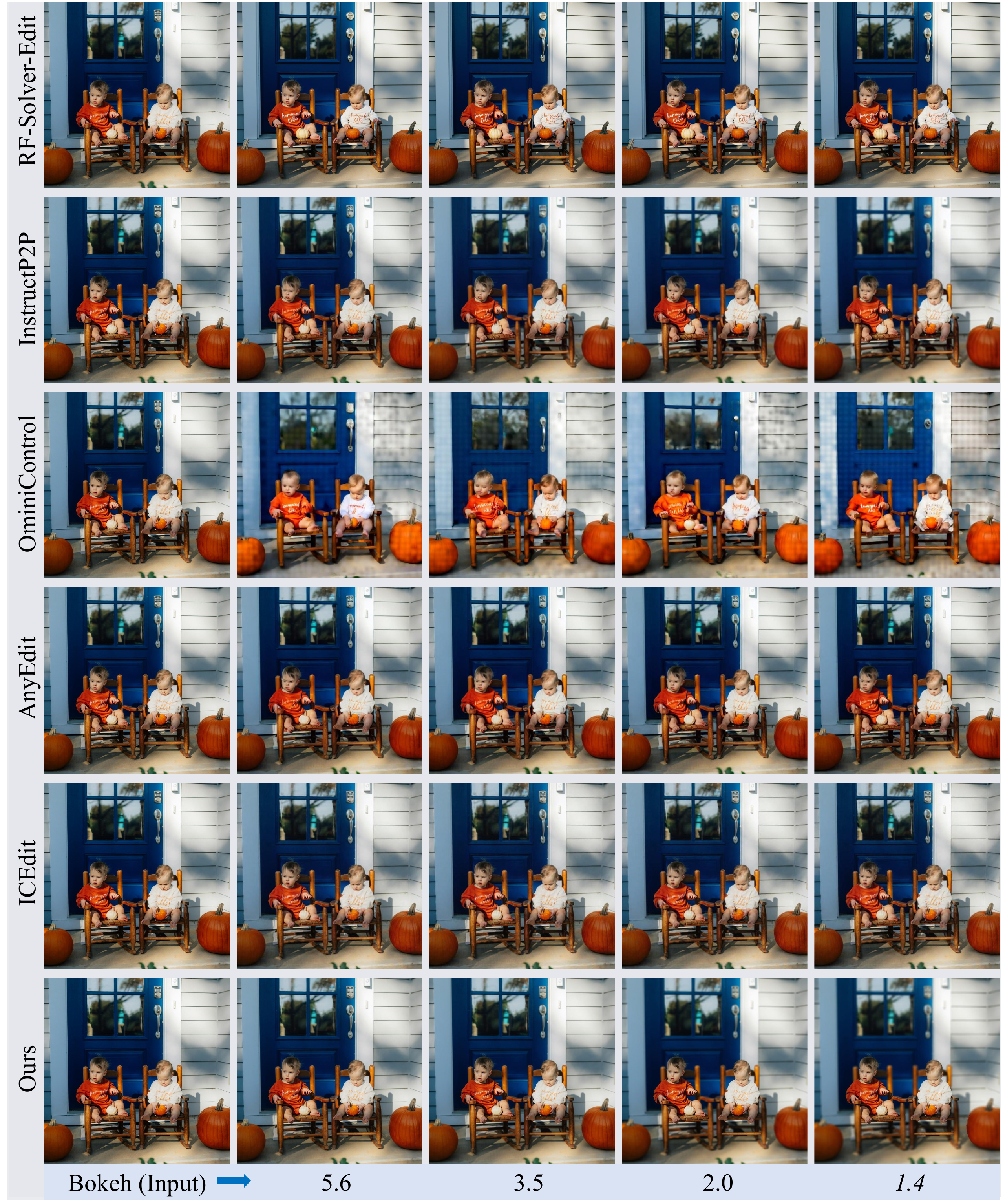}}  
    \vspace{-0.5cm}
    \caption{Visual comparison of \textbf{Bokeh} task retouching results across different methods.}
    \vspace{-0.3cm}
    \label{fig:bokeh}
\end{figure*}

\begin{figure*}[t]
    \centering	  
    \vspace{-0.2cm}
    \centering{\includegraphics[width=1\textwidth]{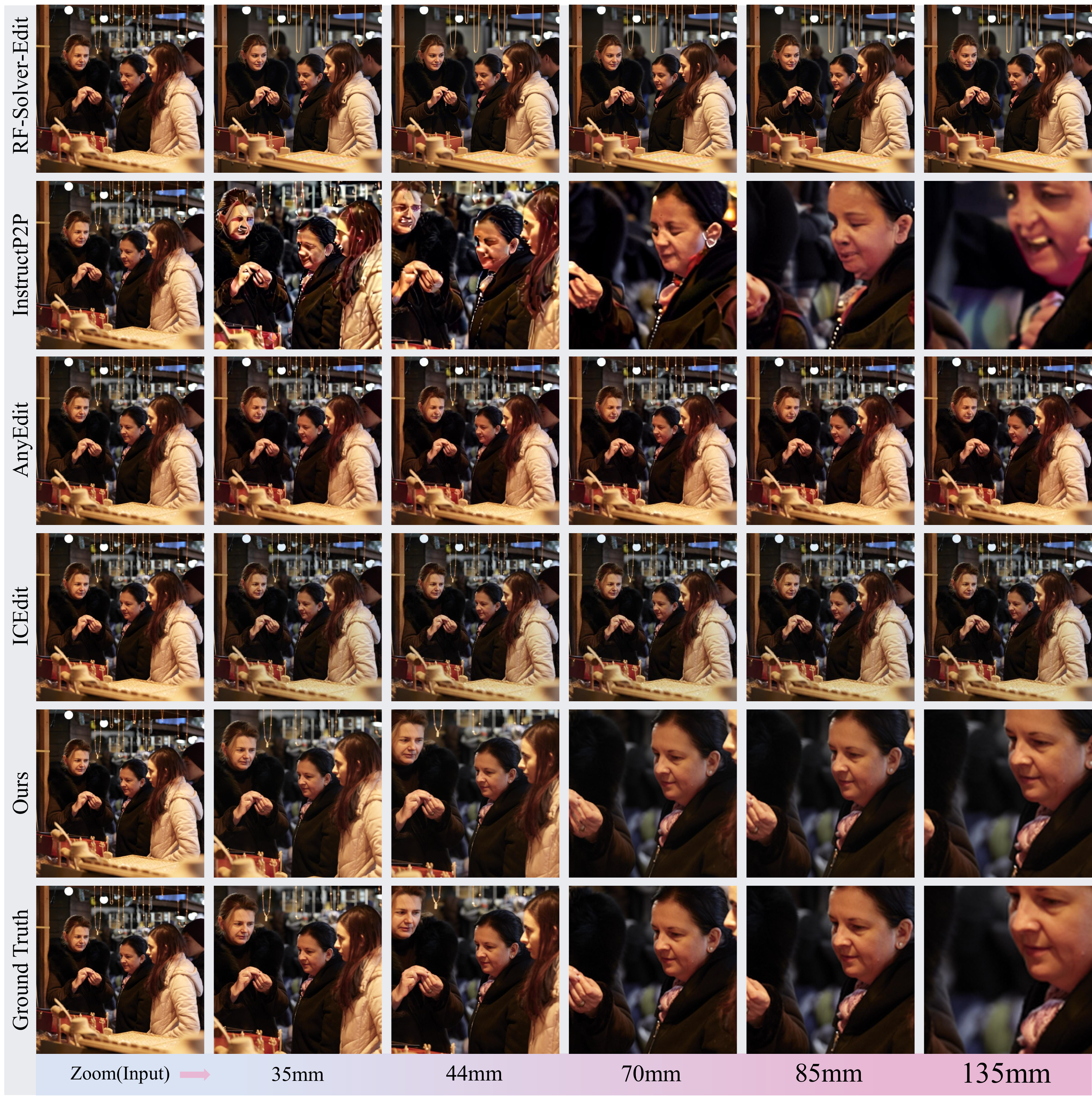}}  
    \vspace{-0.5cm}
    \caption{Visual comparison of \textbf{Zoom} task retouching results across different methods.}
    \vspace{-0.3cm}
    \label{fig:zoom}
\end{figure*}

We provide additional qualitative comparisons on the CameraSet and Real-world datasets.
Fig.~\ref{fig:saturation} to Fig.\ref{fig:zoom} present additional visual comparisons on six photographic retouching tasks: saturation, contrast, film style, exposure, bokeh, and zoom. 
For each task, we gradually vary the corresponding camera parameter and generate a sequence of outputs. 
Across all tasks, CameraMaster produces smooth, monotonic, and visually coherent changes in retouching strength, while preserving scene structure, local details, and subject identity. 
In contrast, compare methods often exhibit non-monotonic behavior, overshooting, or failures to the target setting, and may introduce color cast, banding, or content distortion. 
These qualitative results further confirm that our semantic–parameter decoupling and joint injection enable precise, interpretable control that balances adherence to the photographic directive with fidelity to the original image.

\begin{figure*}[t]
    \centering	  
    \vspace{-0.2cm}
    \centering{\includegraphics[width=1\textwidth]{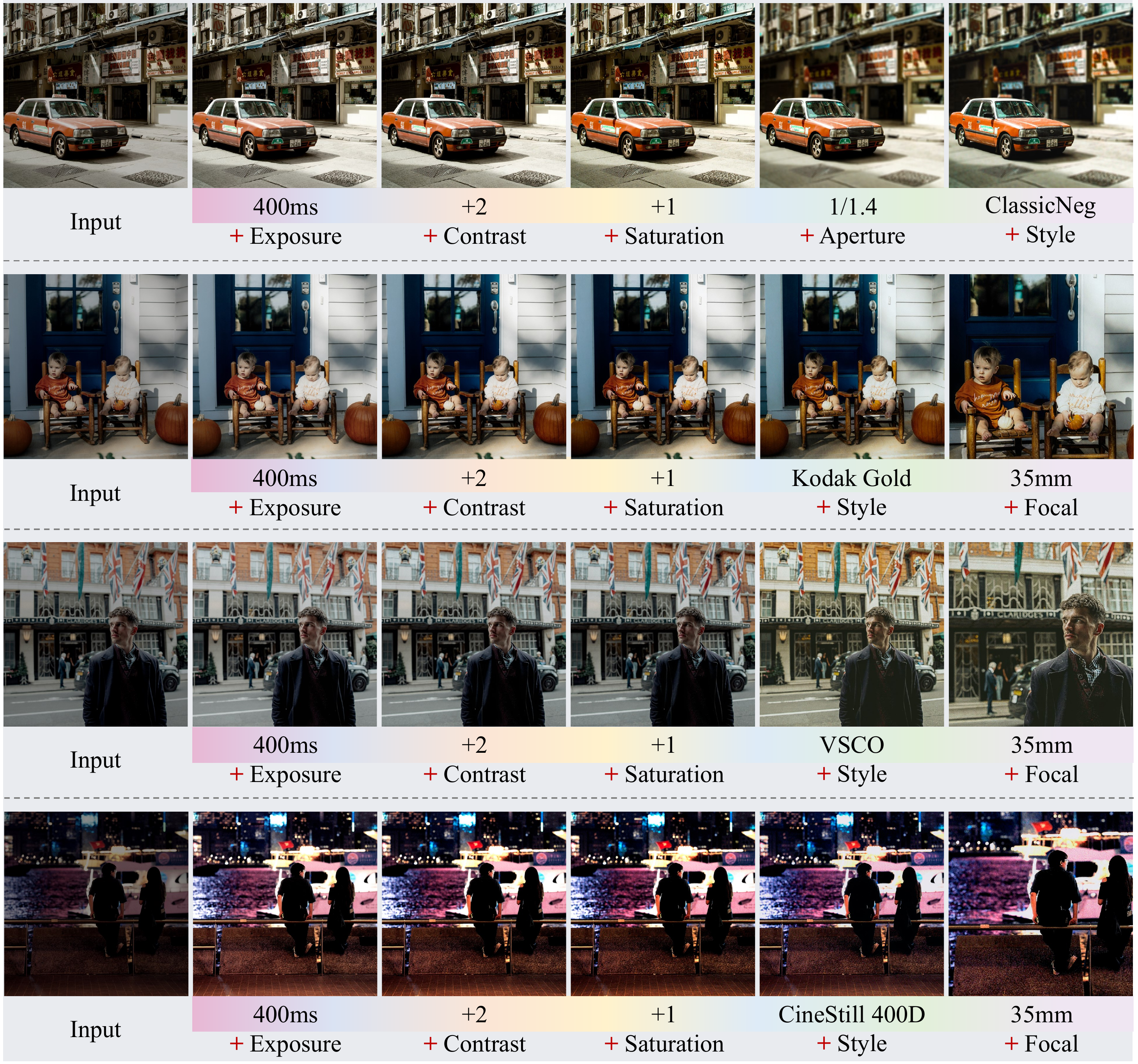}}  
    \vspace{-0.5cm}
    \caption{Examples showcase the \textbf{cross-parameter} retouching power of CameraMaster on the real-world dataset. Our method enables continuous and fine-grained retouching across multiple parameter combinations, demonstrating CameraMaster's universal capability in achieving diverse retouching objectives.}
    \vspace{-0.3cm}
    \label{fig:mulstep}
\end{figure*}

Fig.~\ref{fig:mulstep} shows four representative groups of CameraMaster results on real-world images with joint control over multiple parameters (e.g., exposure, saturation, contrast, and zoom). 
Our model supports continuous and fine-grained transitions across diverse parameter combinations, yielding natural and consistent retouching effects while maintaining stable geometry and identity. 
This demonstrates the generality of CameraMaster as a unified camera-centric interface that can realize diverse retouching goals through intuitive, multi-parameter control on real photographic data.

\vspace{-0.1cm}
\section{Effect of content descriptions.}
\vspace{-0.1cm}
We further study the role of image content descriptions (image captions), as reported in Tab.~\ref{tab:caption}. 
Removing content descriptions and conditioning only on photographic instructions and camera parameters (\emph{w/o Caption}) leads to a consistent degradation across all metrics: PSNR drops from 32.80 to 29.34, SSIM from 0.9240 to 0.9092, while LPIPS and $\Delta E$ increase from 0.0669 to 0.0809 and from 3.073 to 4.126, respectively. 
This shows that content descriptions provide valuable semantic context, enabling better preservation of scene structure and appearance.

At the same time, qualitative results confirm that even without captions, CameraMaster still produces monotonic and consistent retouching along the camera-parameter axis and maintains strong identity preservation. 
In other words, captions are beneficial but not strictly required: they enhance fidelity and perceptual quality, while the core semantic–parameter alignment remains effective in delivering stable, camera-aware control.

\begin{table}[t]
	\centering
	\caption{\textbf{Ablation study on content descriptions (image caption).} We evaluate the performance on the CameraSet test set.}
    \label{tab:caption}
	\vspace{-2mm}
\centering
\small
\renewcommand\arraystretch{1.1}
\resizebox{0.9\linewidth}{!}{
\begin{tabular}{ccccc}
\bottomrule[1pt]\rowcolor[HTML]{FAFAFA}
\multicolumn{1}{c}{Settings}   & PSNR$\uparrow$ & SSIM$\uparrow$ & LPIPS$\downarrow$ & $\Delta \mathbf{E}$ $\downarrow$\\ 
\toprule[0.8pt]
\rowcolor[HTML]{F9FFF9} w/o Caption  &29.34   &0.9092  & 0.0809 & 4.126         \\
\midrule[0.8pt]
\rowcolor[HTML]{F9FBFF} \multicolumn{1}{c}{\ \ \ \
 Full model  } & \textbf{32.80}   & \textbf{0.9240} & \textbf{0.0669} & \textbf{3.073}   \\
\toprule[1pt]
\end{tabular}}
\end{table}

\section{Limitations and Discussion}
\label{limit}
Despite its promising results, CameraMaster has several limitations. 
First, the current implementation supports a maximum resolution of $1024 \times 1024$, which restricts its use in high-resolution professional workflows; naive tiling may break global consistency, and scaling to larger resolutions remains future work. 
Second, although we validate CameraMaster on eight retouching tasks with 38 control levels, the covered functionality is still limited: advanced operations such as shadow, HDR, tone mapping, and relighting are not yet modeled.
In addition, our formulation also targets still-image retouching and does not explicitly address temporal consistency in videos or fine-grained, region-specific edits. 
Finally, the diffusion-based backbone incurs non-trivial computational cost, which can be a bottleneck for latency-sensitive or resource-constrained applications. 
We leave higher-resolution support, richer control spaces, improved generalization, and more efficient inference as important directions for future work.

\end{document}